\documentclass{article}   

\usepackage{arxiv}

\usepackage{graphicx}
\usepackage[colorinlistoftodos]{todonotes}
\usepackage[utf8x]{inputenc}
\usepackage{geometry}
\usepackage{multicol}
\usepackage{amsmath}
\usepackage{url}
\graphicspath{{figs/}}

\usepackage{mathptmx}      
\newif\ifdraft
 \drafttrue
\newcommand{\h}[1]{#1}
\newcommand{\hf}[1]{#1}

\newcommand{\oursystem}{VISIONE}		
\newcommand{\ansearch}{Annotation Search}
\newcommand{\bbsearch}{BBox Search}	
\newcommand{\ocsearch}{OClass Search}
\newcommand{\simsearch}{Similarity Search}
\newcommand{\triplet}{$R_{BB}$-$R_{AN}$-$R_{OC}$}
\newcommand{\rankers}{rankers}

\title{The VISIONE Video Search System:\\  Exploiting Off-the-Shelf Text Search Engines for Large-Scale Video Retrieval}

\author{Giuseppe Amato\\
	ISTI-CNR\\
	Pisa, Italy\\
	\texttt{giuseppe.amato@isti.cnr.it}\And
	Paolo Bolettieri\\
	ISTI-CNR\\
	Pisa, Italy\\
	\texttt{paolo.bolettieri@isti.cnr.it} \And 
	Fabio Carrara\\
	ISTI-CNR\\
	Pisa, Italy\\
	\texttt{fabio.carrara@isti.cnr.it} \And
	Franca Debole\\
	ISTI-CNR\\
	Pisa, Italy\\
	\texttt{franca.debole@isti.cnr.it} \And
	Fabrizio Falchi\\
	ISTI-CNR\\
	Pisa, Italy\\
	\texttt{fabrizio.falchi@isti.cnr.it} \And
	Claudio Gennaro\\
	ISTI-CNR\\
	Pisa, Italy\\
	\texttt{claudio.gennaro@isti.cnr.it} \And
	Lucia Vadicamo\\
	ISTI-CNR\\
	Pisa, Italy\\
	\texttt{lucia.vadicamo@isti.cnr.it} \And
	Claudio Vairo\\
	ISTI-CNR\\
	Pisa, Italy\\
	\texttt{claudio.vairo@isti.cnr.it}
}

\begin{document}
	
	\maketitle

	\begin{abstract}
In this paper, we describe in details VISIONE, a video search system that 
allows users to search for videos using textual keywords, occurrence of objects and their spatial relationships, occurrence of colors and their spatial relationships, and image similarity. These modalities can be combined together to express complex queries and satisfy user needs.
The peculiarity of our approach is that we encode all the information extracted from the keyframes, such as visual deep features, tags, color and object locations, using a convenient textual encoding indexed in a single text retrieval engine. This offers great flexibility when results corresponding to various parts of the query (visual, text and locations) have to be merged.
In addition, we report an extensive analysis of the system retrieval performance, using the query logs generated during the Video Browser Showdown (VBS) 2019 competition. This allowed us to fine-tune the system by choosing the optimal parameters and strategies among the ones that we tested.
		
	\end{abstract}
	
	\keywords{Content-based Video Retrieval  \and Surrogate Text Representation \and Known Item Search \and Ad-hoc Video Search \and Information Search and Retrieval\and  Multimedia and multimodal retrieval\and Video search\and Image search\and Users and interactive retrieval \and Query log analysis}

\section{Introduction}\label{sec:intro}
With the pervasive use of digital cameras and social media platforms, we witness a massive daily production of multimedia content, especially videos and photos. This phenomenon poses several challenges for the management and search of visual archives. On one hand, the use of content-based retrieval systems and automatic data analysis is crucial to deal with visual data that typically are poorly-annotated (think for example to user-generated content). \h{On the other hand}, there is an increasing need of scalable systems and algorithms in order to manage larger data collections.

In this work, we present a video search system, called {\oursystem}, which provides users with various functionalities to easily search for target videos. It relies on artificial intelligence techniques to automatically analyze and annotate visual content and it employs an efficient and scalable search engine to index and search the video data.
A demo of {\oursystem} running on the VBS V3C1 dataset described in the following, is publicly available at \url{http://visione.isti.cnr.it/}.

\begin{figure}[tb]
	\centering
		{\includegraphics[ trim= 0mm 80mm 0mm 0mm,clip,width=1\columnwidth]{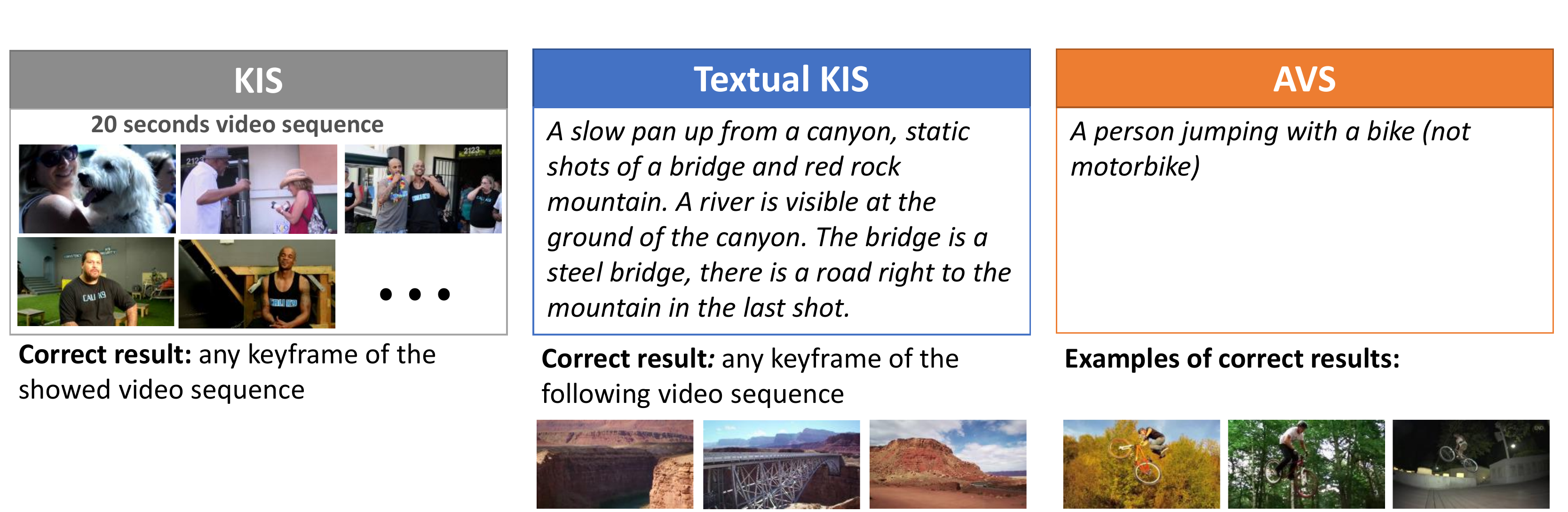}} %
	\caption[]{Examples of KIS, textual KIS and AVS tasks.} 
 \label{fig:VBStask}
\end{figure} 
\h{VISIONE participated at the Video Browser Showdown (VBS) 2019 challenge \cite{VBS2019}.} VBS is an international video search competition \cite{vbs1,vbs2015-2017,VBS2019} that evaluates the performance of interactive video retrieval systems. Performed annually since 2012, it is becoming increasingly challenging as its video archive grows and new query tasks are introduced in the competition. The V3C1 dataset \cite{v3c1} used in the competition since 2019, consists of 7,475 videos gathered from the web, for a total of about 1,000 hours. \h{The V3C1 dataset is segmented into 1,082,657 non-overlapping video segments, based on the visual content of the videos \cite{v3c1}. The shot segmentation for each video as well as the keyframes and thumbnails per video segment are available within the dataset\footnote{\h{\url{https://www-nlpir.nist.gov/projects/tv2019/data.html}}}. In our work, we used the video segmentation and the keyframes provided with the V3C1 dataset.}
\h{The tasks evaluated during the competition are:} 
\textit{Known-Item-Search (KIS)}, \textit{textual KIS} and \textit{Ad-hoc Video Search (AVS)}. Figure \ref{fig:VBStask} gives an example of each task.
The KIS task models the situation in which someone wants to find a particular video clip visually presented, assuming that it is contained in a specific collection of data. The textual KIS is a variation of the KIS task, where the target video clip is no longer visually presented to the participants of the challenge but it is rather described in details by text. This task simulates situations in which a user wants to find a particular video clip, without having seen it before, but knowing the content of the video exactly.
For the AVS task, instead, a general textual description is provided 
and participants need to find as many correct examples as possible, i.e. video shots that fit the given description. 

{\oursystem} can be used to solve both Known-Item and Ad-hoc Video Search tasks. It integrates several content-based analysis and retrieval modules, including a keyword search, a spatial object-based search, a spatial color-based search, and a visual similarity search.
The main novelty of our system is that we specifically designed a textual encoding to be used for indexing and searching video content. This aspect of our system is crucial: we can exploit the latest text search engine technologies, which nowadays are characterized by high efficiency and scalability, without the need to define a dedicated data structure or even worry about implementation issues like software maintenance \h{or} updates to new hardware technologies, etc.

\h{In \cite{Amato2019VBS} we firstly introduced {\oursystem} by listing its functionalities and briefly describing the techniques it employs. In this work, instead, we have two goals: first, to provide a more detailed description of all the functionalities included in {\oursystem} and how each of them are implemented; second, to present an analysis of the system retrieval performance, by examining the logs acquired during the VBS 2019 challenge. Therefore, this manuscript is composed of two parts: firstly we present how all the aforementioned search functionalities are implemented and integrated into a unified framework that relies on a full-text search engine, such as Apache Lucene\footnote{https://lucene.apache.org/}, secondly for the proposed textual encoding, we experimentally studied which text scoring function (ranker) is the best suited for the video search task.}

The rest of the paper is organized as follows. The next section reviews related works. Section \ref{sec:system_description} gives an overview of our system and its functionalities. Key notions on our proposed textual encoding and other aspects regarding the indexing and search phases are presented in Section \ref{sec4}. Section \ref{sec:exp} presents an experimental evaluation to determine which text scoring function is the best in the context of a Know-Item search task. 
Section \ref{sec:conclusion} draws the conclusions.

\section{Related Work}
Video search is a challenging problem of great interest in the multimedia retrieval community. It employs various information retrieval and extraction techniques, such as content-based image and text retrieval, computer vision, speech and sound recognition, and so on. 
\h{
In this context, several approaches for cross-modal retrieval between visual data and text description have been proposed, such as \cite{hu2019scalable,liu2019use,mithun2018learning,otani2016learning,zhen2019deep}, 
to name but a few. Many of them are image-text retrieval methods that make use of a projection of the image features and the text features into the same space (visual, textual or a joint space) so that the retrieval is then performed by searching in this latent space (e.g., \cite{frome2013devise,kiros2014unifying,karpathy2014deep}). Other approaches are referred as video-text retrieval methods as they learn embeddings of video and text in the same space by using different multi-modal features (like visual cues, video dynamics, audio inputs, and text)  \cite{dong2016word2visualvec,miech2019howto100m,mithun2018learning,otani2016learning,pan2016jointly,xu2015jointly}. 
For example, \cite{mithun2018learning} simultaneously utilizes multi-modal features to learn two joint video-text embedding networks: one learns a joint space between text features and visual appearance features, the other learns a joint space between text feature and a combination of activity and audio features.
}

\h{Many video retrieval systems are designed in order to supports complex human generated queries that may include but are not limited to keywords or natural language sentences. Most of them are interactive tools where the users can dynamically refine their queries in order to better specify their search intent during the search process.} 
The Video Browser Showdown (VBS) contest provides a live and fair performance assessment of \h{interactive} video retrieval systems and therefore in recent years has become a reference point for comparing state-of-the-art video search tools. During the competition, the participants have to perform various KIS and AVS tasks in a limited amount of time (generally within 5-8 minutes for each task). To evaluate the interactive search performance of each video retrieval system, several search sessions are performed by involving both expert and novice users\footnote{Expert users are the developers of the in race retrieval system or people that already know and use the system before the competition. Novices are users who interact with the search system for the first time during the competition.}.

Several video retrieval systems participated at the VBS in the last years \cite{vbs2015-2017,VBS2019,VBS2012-2019,VBS2018}. Most of them, including our system, support multimodal search with interactive query formulation. The various systems differ mainly on (i) the search functionalities supported (e.g. query-by-keyword, query-by-example, query-by-sketch, etc.), (ii) the data indexing and search mechanisms used at the core of the system, (iii) the techniques employed during video preprocessing to automatically annotate selected keyframes and extract image features, (iv) the functionalities integrated into the user interface, including advanced visualization and relevance feedback. Among all the systems that participated in VBS, we recall VIRET \cite{VIRET2018}, vitrivr \cite{vitrivr2019}, and SOM-Hunter \cite{SOMHunter2020}, which won the competition in 2018, 2019, and 2020, respectively. 

VIRET \cite{VIRET2018,VIRET2020} is an interactive frame-based video retrieval system that \h{currently} provides \h{four} main retrieval modules (query by keyword, \h{query by free-form text,} queries by color sketch, and query by example). The keyword search relies on automatic annotation of video keyframes. In the latest versions of the system, the annotation is performed using a retrained deep CNN (NasNet \cite{zoph2018learning}) with a custom set of 1243 class labels. A retrained NasNet is also used to extract deep features of the images, which are then employed for similarity search. 
\h{The free-form text search is implemented by using a variant of the W2VV++ model \cite{li2019w2vv++}.} 
An interesting functionality supported by VIRET is the temporal sequence search, which allows a user to describe more than one frame of a target video sequence by also specifying the expected temporal ordering of the searched frames.

Vitrivr \cite{vitrivr2020} is an open-source multimedia retrieval system that supports content-based retrieval of several media types (images, audio, 3D data, and video). For video retrieval, it offers different query modes, including query by sketch (both visual and semantic), query by keywords (concept labels), object instance search, speech transcription search, and similarity search. For the query by sketch and query by example, vitrivr uses several low-level image features and a DNN pixel-wise semantic annotator \cite{rossetto2019query}. The textual search is based on scene-wise descriptions, structured metadata, OCR, and ASR data extracted from the videos. Faster-RCNN \cite{ren2015faster} (pre-trained on the Openimages V4 dataset) and a ResNet-50 [7] (pre-trained on ImageNet) are used to support object instance search. The latest version of vitrivr also supports temporal queries.

SOM-Hunter \cite{SOMHunter2020} is an open-source video retrieval system that supports \h{keyword search, free-text search, and temporal search functionalities, which are implemented as in the VIRET system}. The main novelty of \h{SOM-Hunter} is that it relies on the user's relevance feedback to dynamically update the search results displayed using self-organizing maps (SOMs).

Our system, like almost all current video retrieval systems, relies on artificial intelligence techniques for automatic video content analysis (including automatic annotation and object recognition). 
Nowadays, content-based image retrieval systems (CBIR) are possible solution to the problem of retrieving and exploring a large volume of images resulting from the exponential growth of accessible image data.
Many of these systems use both visual and textual features of the images, but often most of the images are not annotated or only partially annotated. Since manual annotation for a large volume of images is impractical, Automatic Image Annotation (AIA) techniques aim to bridge this gap. For the most part, AIA approaches are based solely on the visual features of the image using different techniques: one of the most common approaches consists in 
training a classifier for each concept and obtaining the annotation results by ranking the class probability~\cite{Chang2003,Carneiro2007}. There are other AIA approaches that aim to improve the quality of image annotation by using the knowledge implicit in a large collection of unstructured text describing images, and are able to label images without having to train a model (Unsupervised Image Annotation approach~\cite{Kobus2001,Li2016,Pellegrin2016}). In particular, the image annotation technique we exploited is an Unsupervised Image Annotation technique originally introduced in \cite{AmatoFGR17}.

Recently, image features built upon Convolutional Neural Networks (CNN) have been used as an effective alternative to descriptors built using image local features, like SIFT, ORB and BRIEF, to name but a few. CNNs have been used to perform several tasks, including image classification, as well as image retrieval \cite{donahue2013decaf,babenko2014neural,razavian2014visual} and object detection \cite{girshick2014rich}. Moreover, it has been proved that the representations learned by CNNs on specific tasks (typically supervised) can be transferred successfully across tasks \cite{donahue2013decaf,razavian2014cnn}.
The activation of neurons of specific layers, in particular the last ones, can be used as features to semantically describe the visual content of an image. Tolias et al. \cite{tolias2015particular} proposed the Regional Maximum Activations of Convolutions (R-MAC) feature representation, which encodes and aggregates several regions of the image in a dense and compact global image representation. Gordo et al. \cite{gordo2016end} inserted the R-MAC feature extractor in an end-to-end differentiable pipeline in order to learn a representation optimized for visual instance retrieval through back-propagation. The whole pipeline is composed by a fully convolutional neural network, a region proposal network, the R-MAC extractor and PCA-like dimensionality reduction layers, and it is trained using a ranking loss based on image triplets. In our work, as a feature extractor for video frames, we used a version of R-MAC that uses the ResNet-101 trained model provided by \cite{gordo2016endArxiv} as the core. This model has proven to perform best on standard benchmarks.

Object detection and recognition techniques also provide valuable information for semantic understanding of images and videos.
In \cite{najva2016sift} the authors propose\h{d} a model for object detection and classification, which integrates Tensor features. The latter are invariant under spatial transformation and together with SIFT features (which are invariant to scaling and rotation) allow improving the classification accuracy of detected objects using a Deep Neural Network.
In \cite{anjum2016video,yaseen2018cloud}, the authors present\h{d} a cloud based system that analyses video streams for object detection and classification. The system is based on a scalable and robust cloud computing platform for performing automated analysis of thousands of recorded video streams. 
The framework requires a human operator to specify the analysis criteria and the duration of video streams to analyze. The streams are then fetched from a cloud storage, decoded and analyzed on the cloud. The framework executes intensive parts of the analysis on GPU-based servers in the cloud. Recently, in \cite{rashid2019object}, the authors proposed an approach that combines Deep Convolutional Neural Network and SIFT. In particular, they extract features from the analyzed images with both approaches, they fuse the features by using a serial-based method that produces a matrix that is fed to ensemble classifier for recognition.

In our system, we used YOLOv3~\cite{yolov3} as CNN architecture to recognize and locate objects in the video frames. The architecture of YOLOv3 jointly performs a regression of the bounding box coordinates and classification for every proposed region. Unlike other techniques, YOLOv3 performs these tasks in an optimized fully-convolutional pipeline that takes pixels as input and outputs both the bounding boxes and their respective proposed categories. This CNN has the great advantage of being particularly fast and at the same time exhibiting remarkable accuracy. To increase the number of categories of recognizable objects, we used three different variants of the same network trained on different data sets, namely, YOLOv3, YOLO9000~\cite{redmon2016yolo9000}, and YOLOv3 OpenImages~\cite{yolov3openimages}.

One of the main peculiarities of our system, compared to others \hf{participating in VBS, is that we decided to employ a full-text search engine to index and search video content, both for the visual  
and textual parts. Since nowadays text search technologies have achieved impressive performance in terms of scalability and efficiency {\oursystem} turns out to be scalable. To take full advantage from these stable search engine technologies, we specifically designed various text encodings for all the features and descriptors extracted from the video keyframes and the user query, and we decided to use the Apache Lucene project.}  
In previous papers, we already exploited the idea of using text encoding, named Surrogate Text Representation \cite{ecdl10}, to index and search image for deep features \cite{amato2018large,AMATO2019IPM,icmr17,ecdl10}. In {\oursystem}, we extend this idea to index also information regarding the position of objects and colors that appear in the images.  

\section{The {\oursystem} video search tool}
\label{sec:system_description}
\begin{figure}[tb]
	\centering
		{\includegraphics[ trim= 30mm 0mm 15mm 0mm,clip,width=1\columnwidth]{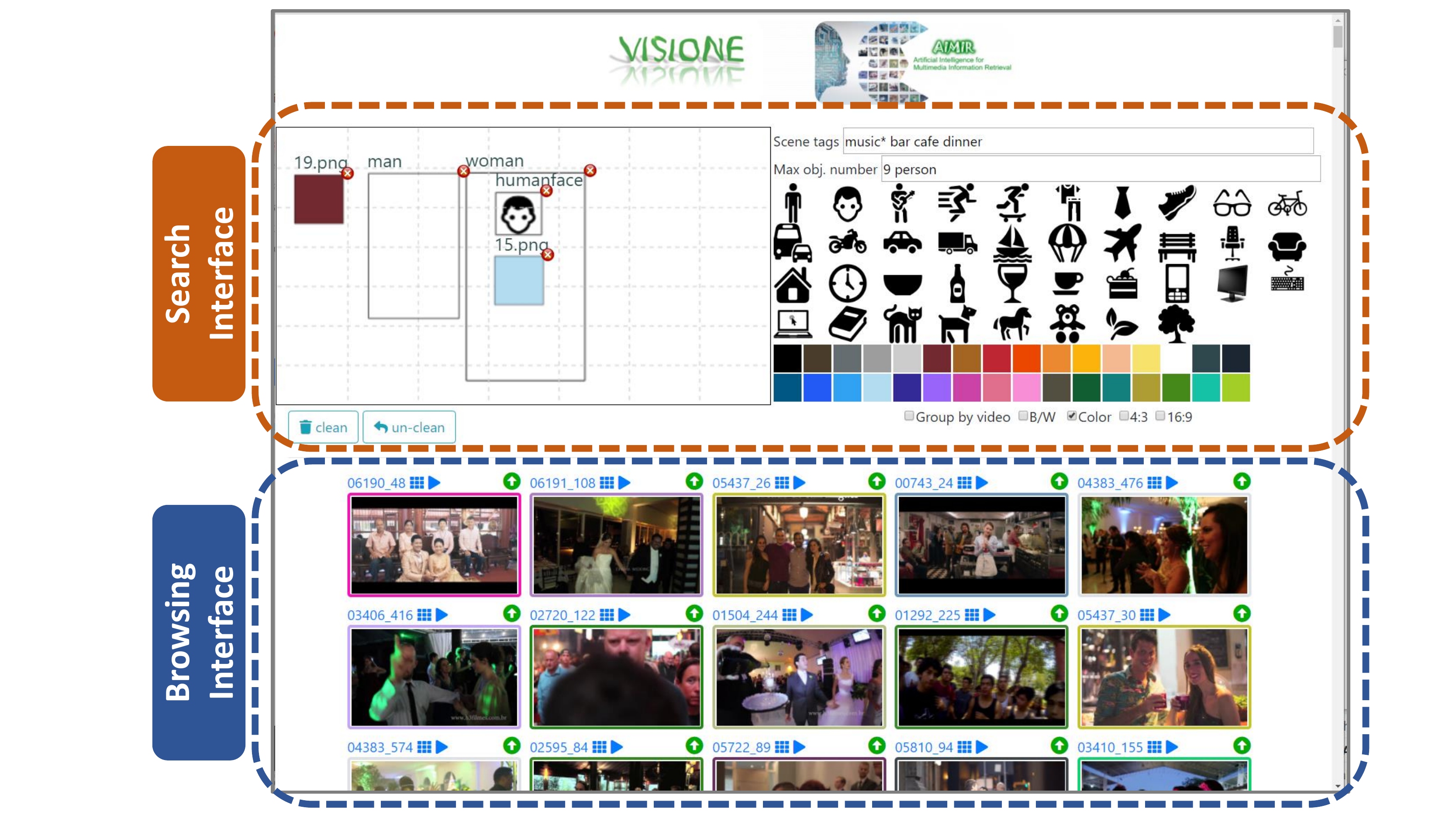}} %
	\caption[]{{\oursystem} User Interface composed by two parts: the search and the browsing.} 
 \label{fig:UI}
\end{figure} 
{\oursystem} is a visual content-based retrieval system designed to support large scale video search.
It allows a user to search for a video \h{describing the content of a scene by formulating textual or visual queries} (see Figure~\ref{fig:UI}). 
 
{\oursystem}, in facts, integrates several search functionalities and exploits deep learning technologies to mitigate the semantic gap between text and image. Specifically it supports:
\begin{itemize}
\item \textit{query by keywords}: the user can specify keywords including scenes, places or concepts (e.g. outdoor, building, sport) to search for video scenes;
\item \textit{query by object location}: the user can draw on a canvas some simple diagrams to specify  the objects that appear in a target scene and their spatial locations;
\item \textit{query by color location}: the user can specify some colors present in a target scene and their spatial locations (similarly to object location above);
\item \textit{query by visual example}: an image can be used as a query to retrieve video scenes that are visually similar to it. 
\end{itemize}
Moreover, the search results can be filtered by indicating whether the keyframes are in color or in b/w, or by specifying its aspect ratio.

\subsection{The User Interface} \label{sec:UI-ST}
The {\oursystem} user interface is designed to be simple, intuitive and easy to use also for users who interact with it for the first time. As shown in Figure ~\ref{fig:UI}, it integrates the \textit{searching} and the \textit{browsing} functionalities in the same window. 

\begin{figure}[!tbp]
	\centering
		{\includegraphics[ trim= 0mm 40mm 5mm 0mm,clip,width=0.97\columnwidth]{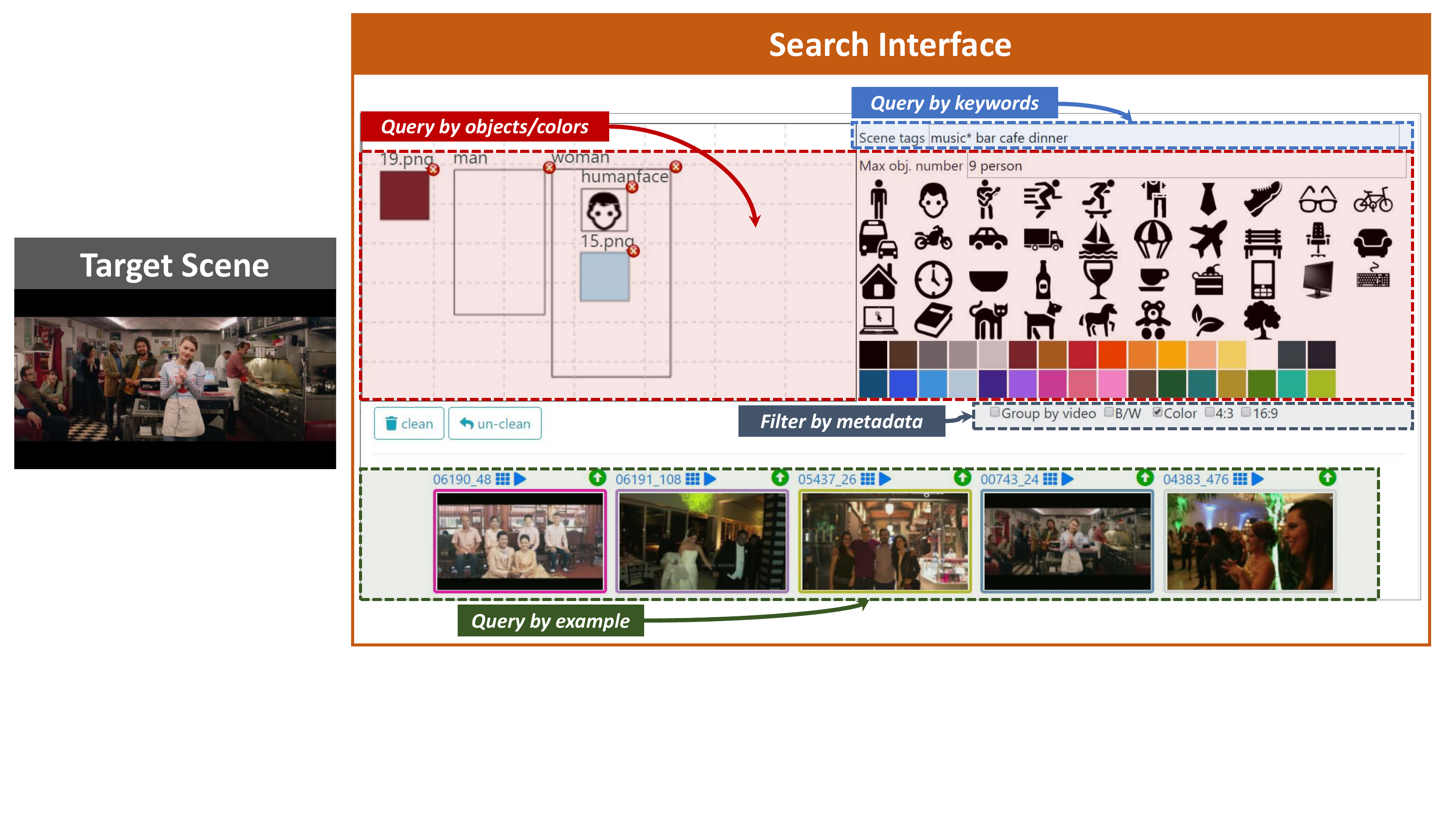}} 
	\caption[]{The video search functionalities as designed in the {\oursystem} User Interface} 
 \label{fig:searchinterface}
\end{figure}

The search interface (Figure~\ref{fig:searchinterface}) provides:
\begin{itemize}
    \item a \textit{text box}, named \textit{``Scene tags"}, where the user can type keywords describing the target scene (e.g. ``park sunset tree walk");
    \item a \textit{color palette} and an \textit{object palette} that can be used to easily drag \& drop a desired color or object on the canvas (see below);
    \item a
    \textit{canvas}, where the user can sketch objects and colors that appear in the target scene simply by drawing bounding-boxes that approximately indicate the positions of the desired objects and colors (both selected from the palettes above) in the scene;
    \item a \textit{text box}, named \textit{``Max obj. number"}, where the user can specify the maximum number of instances of the objects appearing in the target scene (e.g.: two glasses); 
    \item two \textit{checkboxes} where the user can filter the type of keyframes to be retrieved (B/W or color images, 4:3 or 16:9 aspect ratio).
\end{itemize}

The canvas is split into a grid of 7$\times$7 cells, where the user can draw the boxes and then move, enlarge, reduce or delete them to refine the search. 
The user can select the desired color from the palette, drag \& drop it on the canvas and then resize or move the corresponding box as desired.
There are two options to insert objects in the canvas: (i) directly draw a box in the canvas using the mouse and then type the name of the object in a dialog box (auto-complete suggestions are shown to the user), (ii) drag \& drop one of the object icon appearing in the object palette on the canvas. For \h{the user's} convenience, a selection of 38 common (frequently used) objects are included in the object palette.

Note that when objects are inserted in the canvas (e.g. a ``person" and a ``car"), then the system filters out all the images not containing the specified \h{objects} (e.g. all the scenes without a person or without a car). However, images with multiple instances of those objects can be returned in the search results (e.g. images with two or three people and one or more cars). The user can use the \textit{``Max obj. number"} text box to specify the maximum number of instances of an object appearing in the target scene. For example by typing \textit{``1 person 3 car 0 dog"} the system returns only images containing at most one person, three cars and no dog. 

The \textit{``Scene tags"} text box provides auto-complete suggestions to the users and for each tag also indicates the number of keyframes in the databases that are annotated with it. For example, by typing \textit{``music"} the system suggests \textit{``music (204775); musician (1374); music hall (290); ..."}, where the numbers indicates how many images in the database are annotated with the corresponding text (e.g. 204775 images for \textit{``music"}, 1374 images for \textit{``musician"}, etcetera). This information can be exploited by the user when formulating the queries. Moreover, the keyword-based search supports wildcard matching. For example, with \textit{``music$*$"} the system searches for any tag that starts with \textit{``music"}.

Every time the user interact with the search interface (e.g type some text or add/ move/delete a bounding box) the system automatically updates the list of search results, which are displayed in the browsing interface, immediately below the search panel. In this way the user can interact with the system and gradually compose his query by also taking into account the search results obtained so far to refine the query itself. 

\begin{figure}[!tb]
	\centering
		{\includegraphics[ trim= 40mm 40mm 40mm 0mm,clip,width=0.97\columnwidth]{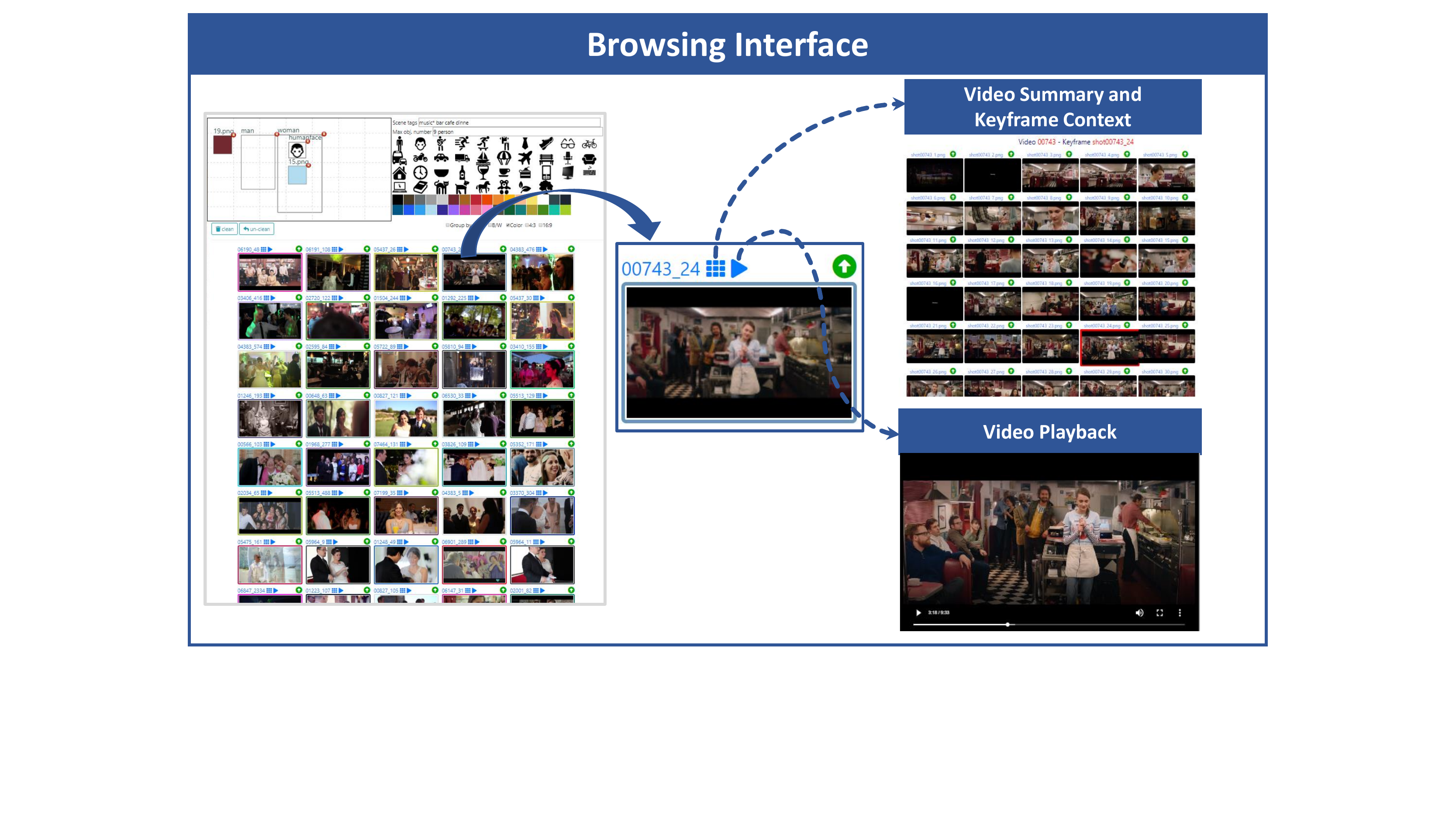}} %
	\caption[]{Browsing Interface: for each keyframe result it allows accessing the information such as video summary, keyframe context, play the video starting from the selected keyframe and search similar keyframe.
}  \label{fig:browsing}
\end{figure}

The browsing part of the user interface (Figure \ref{fig:browsing}) allows accessing the information associated with the video, every displayed keyframe belongs to \h{it}, a keyframe-based video summary and \h{playing} the video starting from the selected keyframe. In this way, the user can easily check if the selected image belong to the searched video. \h{The search results can also be grouped together according to the fact that the keyframes belong to the same video. This visualization option can be enabled/disabled by clicking on the \textit{``Group by video"} checkbox. Moreover, while browsing the results, the user can use one of the displayed image to perform an image {\simsearch} and retrieve frames visually similar to the one selected. A {\simsearch} is executed by double clicking on an image displayed in the search results.}  

\subsection{System Architecture Overview}
\label{sec:IN-ST}

\begin{figure}[!tb]
	\centering
		{\includegraphics[ trim= 10mm 100mm 10mm 5mm,clip,width=\columnwidth]{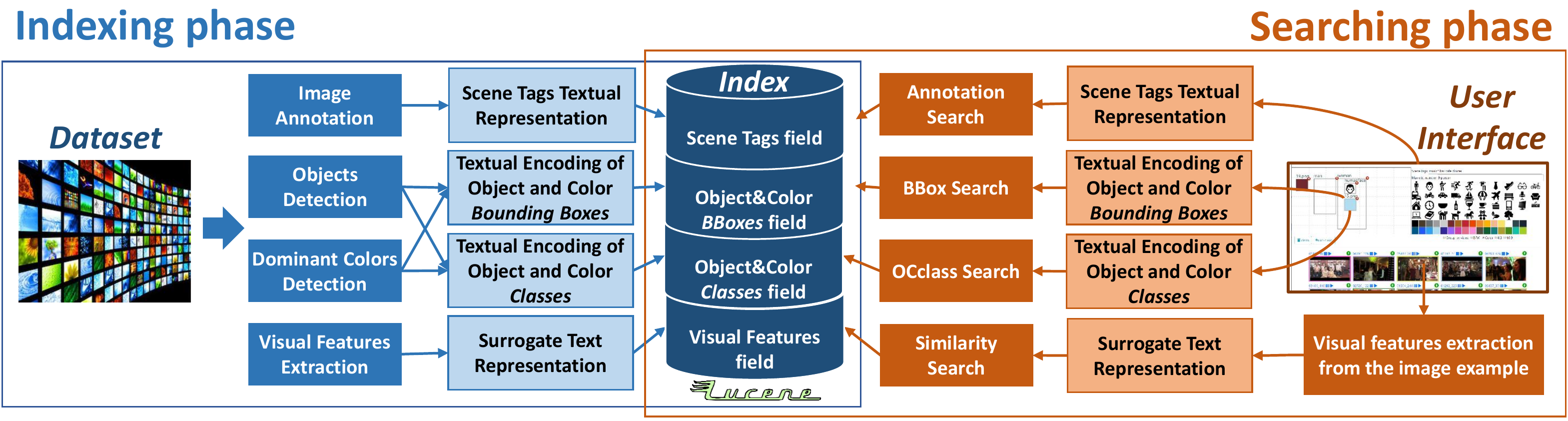}} %
	\caption[]{System Architecture: a general overview of the components of the two main phases of the system, the indexing and the browsing.} 
 \label{fig:architecture}
\end{figure}

The general architecture of our system is illustrated in Figure \ref{fig:architecture}. Each component of the system will be described in detail in the following sections; here we give an overview of how it works.
To support the search functionalities introduced above, our system exploits deep learning technologies to understand and represent the visual content of the database videos. Specifically, it employs:
\begin{itemize}
    \item an image annotation engine, to extract scene tags (see Sec. \ref{sec:Annotation});
    \item  state-of-the-art object detectors, like YOLO \footnote{https://pjreddie.com/darknet/yolo/}, to identify and localize objects in the video keyframes (see Sec \ref{sec:ObjectColorDetection});
    \item spatial colors histograms, to identify dominant colors and their locations (see Sec \ref{sec:ObjectColorDetection});
     \item the R-MAC \cite{tolias2015particular} deep visual descriptors, to support the {\simsearch} functionality (see Sec. \ref{sec:SimilaritySearch})
\end{itemize}

The peculiarity of the approach used in {\oursystem} is to represent all the different types of descriptors extracted from the keyframes (visual features, scene tags, colors/object locations) with a textual encoding that is indexed in a single text search engine. 
This choice allows us to exploit mature and scalable full-text search technologies and platforms for indexing and searching \h{video repository}. 
In particular, {\oursystem} relies on the Apache Lucene full-text search engine. The text encoding used to represent the various types of information, associated with every keyframe, is discussed in Section \ref{sec4}.

Also the queries formulated by the user through the search interface (e.g. the keywords describing the target scene and/or the diagrams depicting objects and the colors locations) are transformed into textual encoding, in order to process them. We designed a specific textual encoding for each typology of data descriptor as well as for the user queries. 

\h{In the full-text search engine,} the information extracted from every keyframe \h{is} composed of four textual fields, as shown in Figure \ref{fig:architecture}:
\begin{itemize}
    \item \textit{Scene Tags}, containing automatically associated tags\h{;}
     \item \textit{Object\&Color BBoxes}, containing text encoding of colors and objects locations\h{;}
     \item \textit{Object\&Color Classes}, containing global information on objects and colors in the keyframe\h{;}
     \item \textit{Visual Features}, containing text encoding of extracted visual features.
\end{itemize}
These four fields are 
used to serve the four main search operations of our system:
\begin{itemize}
    \item \textit{{\ansearch}}, search for keyframes associated with specified annotations\h{;}
    \item \textit{{\bbsearch}}, search for keyframes having specific spatial relationships among objects/colors\h{;}
    \item \textit{{\ocsearch}}, search for keyframes containing specified objects/colors\h{;}
    \item \textit{{\simsearch}}, search for keyframes visually similar to a query image
\end{itemize}
The user query is broken down into four sub-queries (one for each search operation), and a query rescorer (the Lucene QueryRescorer implementation in our case) is used to combine the search results of all the sub-queries. 
\h{In the next section, we will describe the four search operations and further details on the indexing and searching phases.}

\section{Indexing and Searching Implementation} \label{sec4}

In {\oursystem}, as already anticipated, content of keyframes is represented and indexed using automatically generated annotations, positions of occurring objects, positions of colors, and deep visual features. In the following we describe how these descriptors are extracted, indexed, and searched.

\subsection{Image Annotation}\label{sec:Annotation}
One of the most natural way of searching in a large multimedia data set is using a keyword-based query. To support such kind of queries, we employed our automatic annotation system\footnote{Demo available at \url{http://mifile.deepfeatures.org}} \h{that is} introduced in \cite{AmatoFGR17}. \h{This system} is based on an unsupervised image annotation approach that exploits the knowledge implicitly existing in a huge collections of unstructured texts describing images\h{, allowing} us to annotate the images without using a specified trained model. The advantage is that the target vocabulary we used for the annotation reflects well the way people actually describe their pictures. Specifically, our system uses the tags and the descriptions contained in the metadata of a large set of media selected from the Yahoo Flickr Creative Commons 100 Million (YFCC100M) dataset \cite{Thomee2016}. Those tags are validated using WordNet \cite{fellbaum98wordnet}, cleaned and then used as the knowledge base for the automatic annotation.

The subset of the YFCC100M dataset that we used for building the knowledge base was selected by identifying images with relevant textual descriptions and tags. To this scope, we used a metadata cleaning algorithm that leverages on the semantic similarities between images. 
Its core idea is that if a tag is contained in the metadata of a group of very similar images, then that tag is likely to be relevant for all these images. The similarity between images was measured by means of visual deep features; specifically, we used the output of the sixth layer of the neural network Hybrid-CNN \footnote{Publicly available in the Caffe Model Zoo, \url{http://github.com/BVLC/caffe/wiki/Model-Zoo}} as visual descriptors. As a result of our metadata cleaning algorithm we selected about 16 thousands terms associated to about one million images. The set of deep features extracted from those images were then indexed using the MI-file index~\cite{Amato2012} in order to allow us to access the data and perform similarity search in a very efficient way.

The annotation engine is based on a k-NN classification algorithm. An image is annotated with the most frequent tags associated with the most similar images in the YFCC100M cleaned subset. The specific definition of the annotation algorithm is out of the scope of this paper and we refer to \cite{AmatoFGR17} for further details. 

\begin{figure}[!tb]
	\centering
		{\includegraphics[ trim= 5mm 5mm 5mm 0mm,clip,width=\columnwidth]{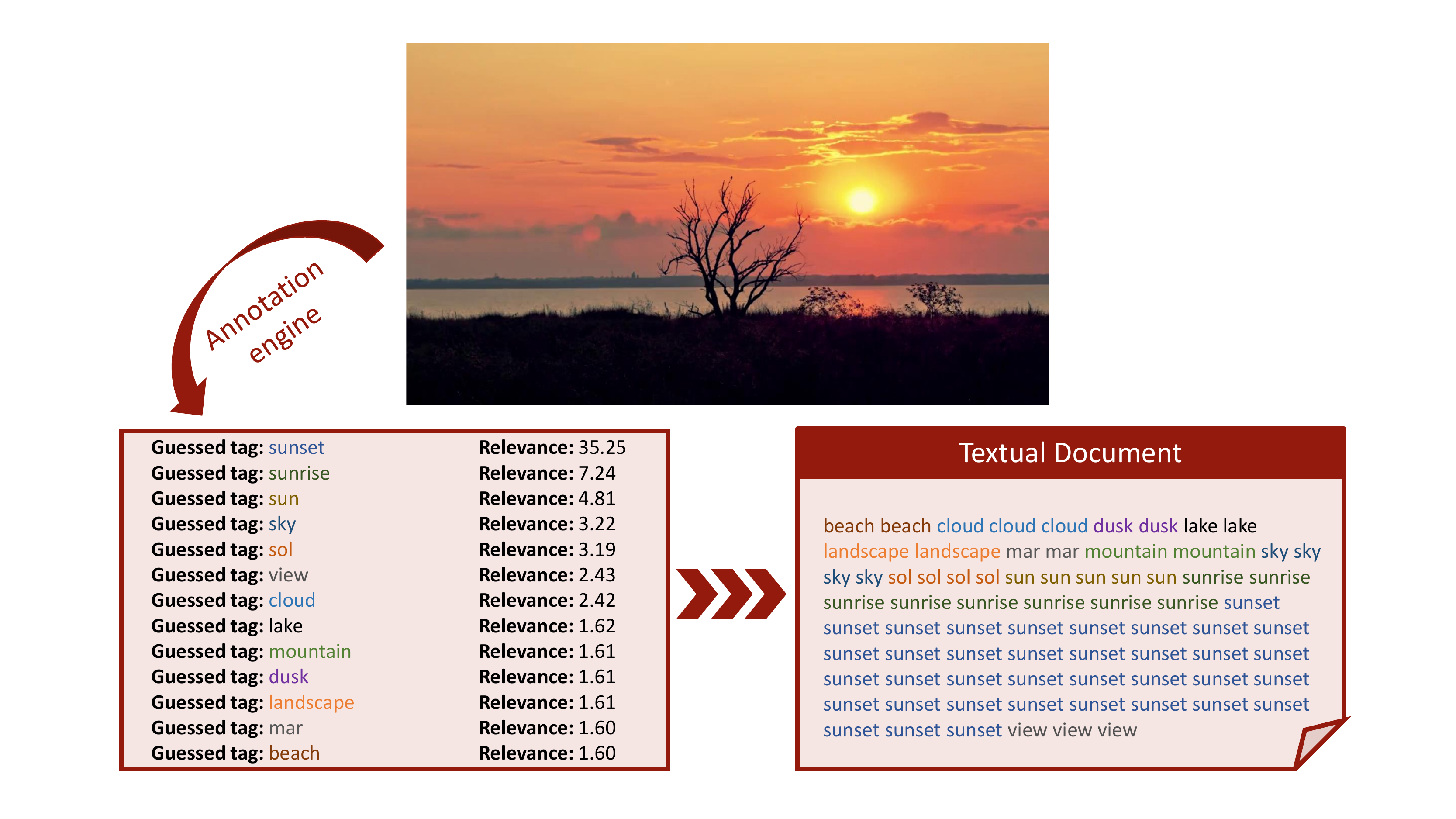}} %
	\caption[]{Example of our image annotation and its representation as single textual document. In the textual document, each tag is repeated a number of time equal to the least integer greater than or equal to the tag relevance.} 
 \label{fig:exampleAnnotation}
\end{figure}

In Figure \ref{fig:exampleAnnotation}, we show an example of annotation obtained with our system. Please note that our system also provide a relevance score to each tag associated to the image. The bigger the score the more relevant the tag.
We used our annotation system to label the video keyframes of the V3C1 dataset. 
For each keyframe we produce a ``tag textual encoding" by concatenating all the tags associated to the images.
In order to represent the relevance of the associated tag, each tag is repeated a number of time equal to the relevance score of the tag itself (the relevance of each tag is approximated to an integer using the ceiling function). The ordering of the tags in the concatenation is not important because what matters are the tag frequencies. In Figure \ref{fig:exampleAnnotation} the box named \textit{Textual Document} shows an example of concatenation associated to a keyframe.

\paragraph{{\ansearch}.}
The annotations, generated as described above, can be used to retrieve videos, by typing keywords in the \textit{``Scene tags"} text box of the user interface (see Figure \ref{fig:searchinterface}). As already anticipated in Section \ref{sec:IN-ST}, we call \textit{{\ansearch}} this searching option. The {\ansearch} is executed performing a full-text search. As described in Section \ref{sec:exp}, during the VBS competition the BM25 similarity was used as text scoring function.

\subsection{Objects and Colors}
\label{sec:ObjectColorDetection}

\begin{figure}[tb]
	\centering
		{\includegraphics[ trim= 45mm 40mm 45mm 5mm,clip,width=\columnwidth]{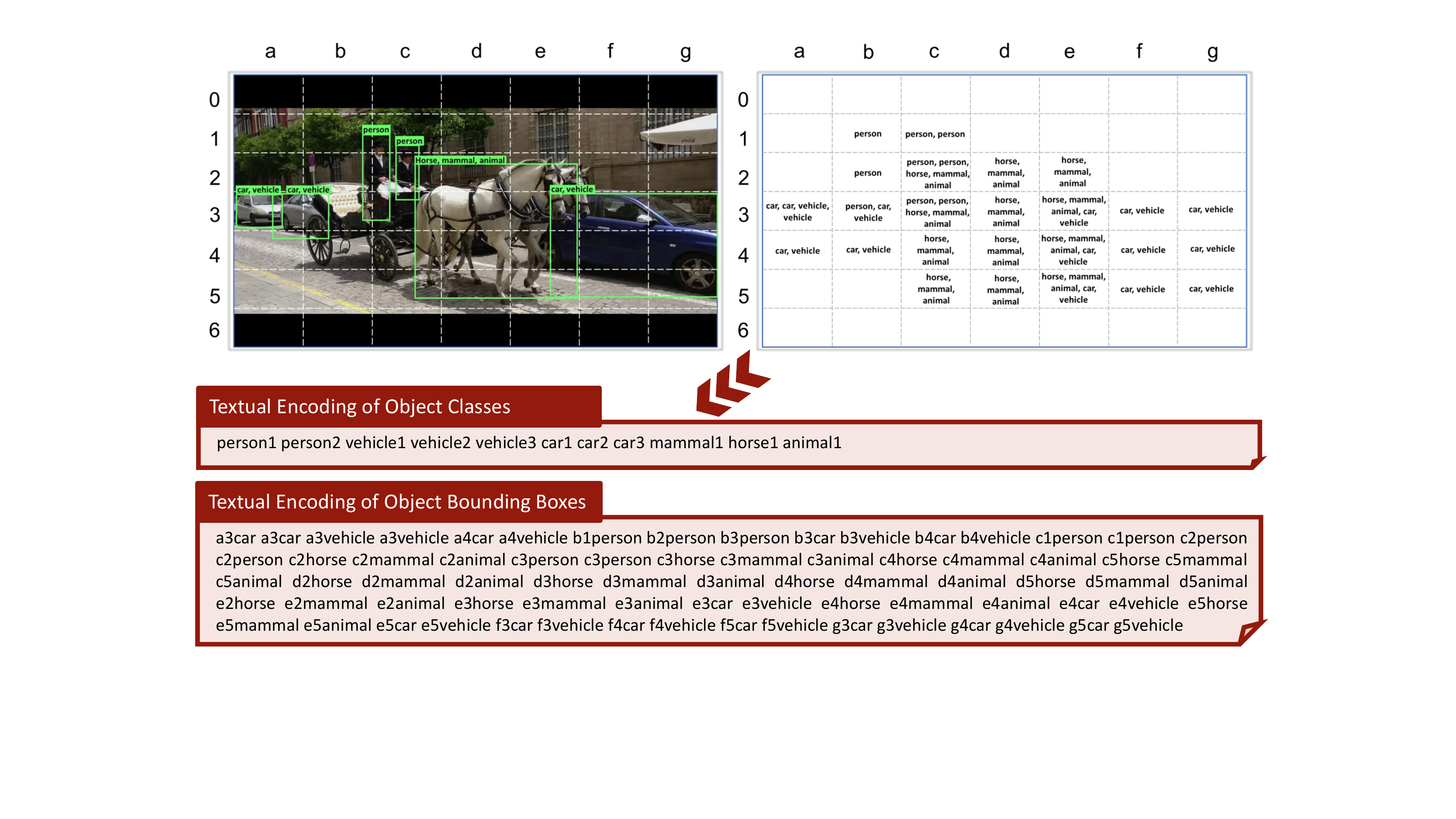}} %
	\caption[]{Example of our textual encoding for objects and their spatial locations. The textual encoding for color locations is obtained in a similar way.} 
 \label{fig:textualencoding}
\end{figure}

Information related to objects and colors in a \h{keyframe} are treated in a similar way in our system. Given a keyframe we store both local and global information about objects and colors contained in it. As we discussed in Section \ref{sec:IN-ST}, the positions where objects and colors occur are stored in the \textit{Object\&Color BBoxes} field; all objects and colors occurring in a frame are stored in the \textit{Object\&Color Classes} field.
\subsubsection{Objects}
We used a combination of three different versions of YOLO to perform object detection: YOLOv3 \cite{yolov3}, YOLO9000 \cite{redmon2016yolo9000}, and YOLOv3 OpenImages \cite{yolov3openimages}, to extend the number of detected objects. 
The idea of using YOLO to detect objects within video has already been exploited in VBS, e.g. by Truong et al. \cite{Truong:VBS2018}. The peculiarity of our approach is that we combine and encode the spatial position of the detected objects in a single textual description of the image. In particular, each object detected in the image \textit{I} is indexed using a specific textual encoding $ENC=(cod_{loc}cod_{class})$ that puts together the location $cod_{loc}$ and the class $cod_{class}$ corresponding to the object. 
To obtain this information, we use a grid of $7 \times 7$ cells overlaid to the image to determine where (over which cells) each object is located. 
The textual encoding of this information is created as follows. For each image, we have a space-separated concatenation of \textit{ENC}s, one for all the cells ($cod_{loc}$) in the grid that contains the object ($cod_{class}$): for example, for the image in Figure \ref{fig:textualencoding} the rightmost car is indexed with the sequence $\{e3$\textit{car} $f3$\textit{car} ... $g5$\textit{car}$\}$ where ``car" is the $cod_{class}$ of the object \textit{car}, $e3$ is the $cod_{loc}$ of the cell at column \textit{``e"} and row $3$, $f4$ is the $cod_{loc}$ of the cell at column \textit{``f"} and row $3$, etc.
This information is stored in the \textit{Object\&Color BBoxes} field of the record associated with the keyframe.
In addition to the position of objects, we also maintain global information about the objects contained in a keyframe, in terms of number of occurrences of each object detected in the image (see Figure \ref{fig:textualencoding}). Occurrences of objects in a keyframe are encoded by repeating the object ($cod_{class}$) as many times as the number of the occurrences ($cod_{occ}$) of the object itself.
This information is stored using an encoding that composes the classes with their occurrences in the image: ($cod_{class}cod_{occ}$).
For example, in Figure \ref{fig:textualencoding}, YOLO detected 2 persons, 3 cars, which are also classified as vehicle by the detector, and 1 horse, also classified as animal and mammal, and this results in the Object Classes encoding as “\textit{person1 person2 vehicle1 vehicle2 vehicle3 car1 car2 car3 mammal1 horse1 animal1}”. This information is stored in the \textit{Object\&Color Classes} field of the record associated with the keyframe.

\subsubsection{Colors}
To represent colors, we use a palette of 32 colors\footnote{https://lospec.com/palette-list} which represents a good trade-off between the huge miscellany of colors and simplicity of choice for the user at search time.
For the creation of the color textual encoding we used the same approach employed to encode the object classes and locations, using the same grid of $7\times 7$ cells.
To assign the colors to each cell of the grid we used the following approach. We first evaluate the color of each pixel by using the CIELAB color space. Then, we map the evaluated color of the pixel to our 32-colors palette. To do so, we perform a $k$-NN similarity search between the evaluated color and our 32 colors to find the colors in our palette that most match the color of the current pixel. The metric used for this search is the Earth Mover's Distance \cite{Rubner97}. 
We take into consideration the first two colors in $k$-NN results. The first color is assigned to that pixel. We then compute the ratio between the scores of the two colors and if it is greater than 0.5 then we also assign the second color to that pixel. This is done to allow matching of very similar colors during searching.
We repeat this for each pixel of a cell in the grid and then we sum the occurrences of each color of our palette for all the pixels in the cell. Finally, we assign to that cell all the colors whose occurrence is greater than 7\% of the number of pixels contained in the cell. So more than one color may be assigned to a single cell. This redundancy helps reducing misclassified colors from what they appear to the human eye.

The colors assigned to all the $7\times 7$ images cells are then encoded into two textual documents, one for the color locations and one for the global color information, using the same approach employed to encode object classes and locations, and discussed above. Specifically, the textual document associated to the color location is obtained by concatenating textual encodings of the form $cod_{loc}cod_{class}$, where $cod_{loc}$ is an identifier of a cell and $cod_{class}$ is the identifier of a color assigned to the cell. This information is stored in the \textit{Object\&Color BBoxes} field. The textual document for the color classes is obtained by concatenating the text identifiers ($cod_{class}$) of all the colors assigned to the image. This information is stored in the \textit{Object\&Color Classes} field of the record associated with the keyframe.
\paragraph{Object and Color Location Search.}
At run-time phase, the search functionalities for both the query by object and color location are implemented using two search operations: the bounding box search (\textit{{\bbsearch}}) and the object/color-class search (\textit{{\ocsearch}}).

The user can draw a bounding box in a specific position of the canvas and specify which object/color wants to found in that position, or he/she can drag \& drop a particular object/color from the palette in the user interface and resize the corresponding bounding box as desired (as shown in the ``Query by object/colors'' of Figure \ref{fig:searchinterface}).
All the bounding boxes present in the canvas, both related to colors and objects, are then converted into the two textual encoding described above (Object\&Color Bounding Boxes and Object\&Color Classes encodings).

Then for the actual search phase, first an instance of the {\ocsearch} operator is executed. This operator tries to find a match between all the objects represented in the canvas and the frames stored in the index that contains these objects. During the VBS2019 competition the metric used to find this match was the dot product.
This produces a result set containing a subset of the dataset with all the frames that match the objects represented in the canvas. 
After this, the {\bbsearch} operator performs a rescoring of the result set by matching the textual encoding of the Object and Color Bounding Boxes encoding of the query 
with all the corresponding encodings in the index. The metric used in this case during the VBS competition was BM25.
After the execution of these two search operators, the frames that satisfied these two searches ordered by descending score are shown in the browsing part of the user interface.

\subsection{Deep Visual Features}\label{sec:SimilaritySearch}

\begin{figure}[!tb]
	\centering
		{\includegraphics[ trim= 5mm 65mm 5mm 0mm,clip,width=\columnwidth]{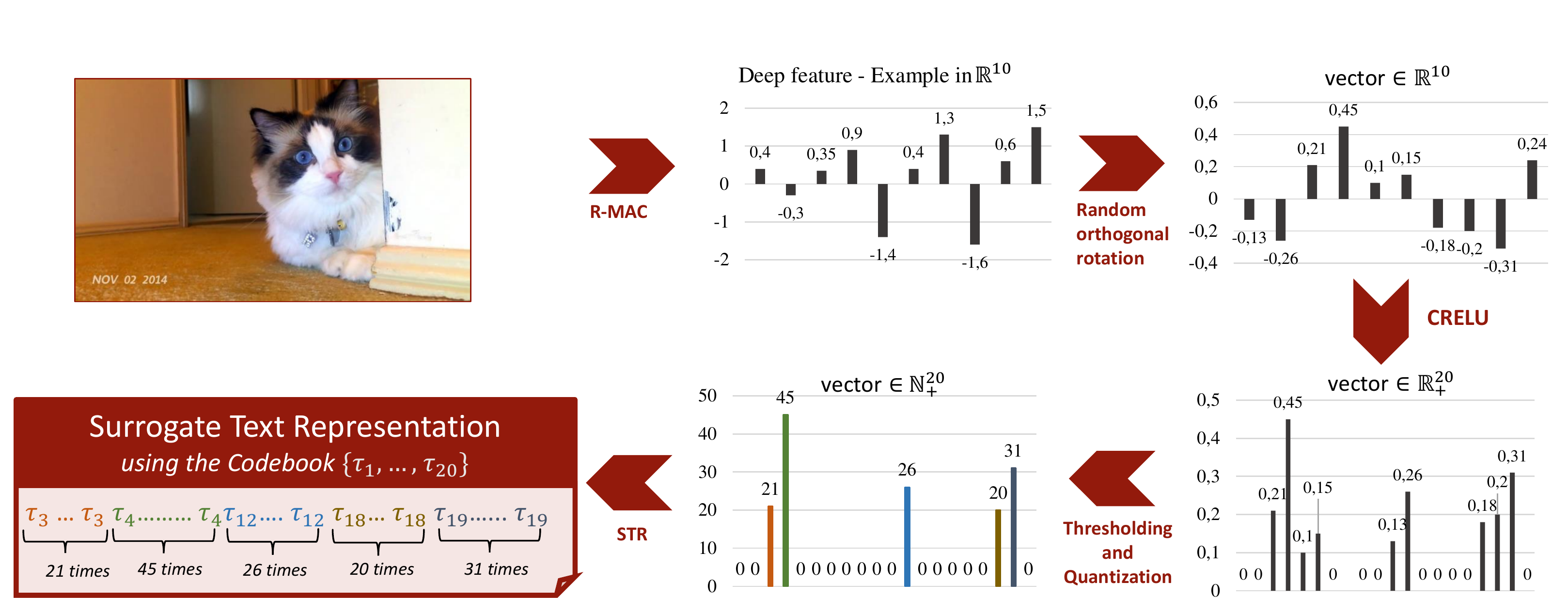}} %
	\caption[]{Scalar Quantization-based Surrogate Text representation: clockwise the transformation of the image's R-MAC descriptor- depicted as a $10$-dimensional vector- into a textual encoding called Surrogate Text Representation.} 
 \label{fig:exampleSTR}
\end{figure}
{\oursystem} also supports content-based visual search functionality, i.e., it allows users to retrieve keyframes visually similar to a query image given by example. 
In order to represent and compare the visual content of the images, we use the Regional Maximum Activations of Convolutions (R-MAC) \cite{tolias2015particular}, which is a state-of-art descriptor for image retrieval. 
The R-MAC descriptor effectively aggregates several local convolutional features (extracted at multiple position and scales) into a dense and compact global image representation. We use the ResNet-101 trained model provided by Gordo et al. \cite{gordo2016end} as an R-MAC feature extractor since it achieved the best performance on standard benchmarks. The used R-MAC descriptors are $2048$-dimensional real-valued vectors.  

To efficiently index the R-MAC descriptor, we transform the deep features into a textual encoding suitable for being indexed by a standard full-text search engine. We used the \textit{Scalar Quantization-based Surrogate Text representation} to transform the deep features into a textual encoding, which was proposed in \cite{AMATO2019IPM}. The idea behind this approach is to map the real-valued vector components of the R-MAC descriptor into a (sparse) integer vector that acts as the term frequencies vector of a synthetic codebook. 
Then the integer vector is transformed into a text document by simply concatenating some synthetic codewords so that the term frequency of the $i$-th codeword is exactly the $i$-th element of the integer vector. For example, the four-dimensional integer vector $[2,1,0,1]$ is encoded with the text ``$\tau_1 \, \tau_1\, \tau_2\, \tau_4$", where $\{\tau_1, \tau_2, \tau_3, \tau_4\}$ is a codebook of four synthetic alphanumeric terms. The overall process used to transform an R-MAC descriptors into a textual encoding is summarized in Figure \ref{fig:exampleSTR} (for simplicity, the R-MAC descriptor is depicted as a $10$-dimensional vector).

The mapping of the deep features into the term frequencies vectors is designed (i) to  preserve as much as possible the rankings, i.e. similar features should be mapped into similar term frequencies vectors (for effectiveness) and (ii) to produce sparse vectors, since each data object will be stored in as many posting lists as the non-zero elements in its term frequencies vector (for efficiency).
To this end, the deep features are first centered using their mean and then rotated using a random orthogonal transformation. The random orthogonal transformation is particularly useful to distribute the variance over all the dimensions of the vector as provide good balancing for high dimensional vectors without the need to search for an optimal balancing transformation.
In this way, we try to increase the cases where the dimensional components of the features vectors have same mean and variance, with mean equal to zero. Moreover the used roto-traslation preserves the rankings according to the dot-product (see \cite{AMATO2019IPM} for more details).
Since search engines, like the one we used, use an inverted file to store the data, as a second step, we have to sparsify the features. Sparsification guarantees the efficiency of these indexes. To achieve this, Scalar Quantization approach maintains components above a certain threshold by zeroing all the others and quantize the non-zero elements to integer values. To deal with negative values the Concatenated Rectified Linear Unit (CReLU) transformation \cite{crelu} is applied before the thresholding. Note that the CReLU simply makes an identical copy of vector elements, negate it, concatenate both original vector and its negation, and then zeros out all the negative values. Eventually, as the last operation, we apply the \textit{Surrogate Text Representation} technique \cite{ecdl10} that allows us to transform an integer vector into text by generating a text document that repeats the codewords associated with the components of the vector a number of times proportional to the value of the components themselves. 

In {\oursystem} the Surrogate Text Representation of a dataset image is stored in the \textit{``Visual Features''} field of our index (Figure \ref{fig:architecture}).

\paragraph{{\simsearch}.}
{\oursystem} relies on the Surrogate text encodings of images to perform the {\simsearch}. 
When the user starts a {\simsearch} by selecting a keyframe in the browsing interface, the system retrieves all the indexed keyframes whose Surrogate Text Representation are similar to the Surrogate Text Representation of the selected keyframe. We used the dot product over the  frequency terms vectors (TF ranker) as text similarity function since it achieved very good performance for large-scale image retrieval task \cite{AMATO2019IPM}.

\subsection{Overview of the Search Process } 
As we described so far, our system relies on four \textit{search operations}: an {\ansearch}, a {{\bbsearch}}, an {{\ocsearch}}, and a {{\simsearch}}.
Every time a user interacts with the {\oursystem} interface (add/remove/update a bounding box, add/remove a keyword, click on an image, etc...), a new query $Q$ is executed, where $Q$ is the sequence of the instances of search operations currently active in the interface.
The query is then split into subqueries, where a \emph{subquery} contains instances of a single search operation. 
In a nutshell, the system runs all the subqueries using the appropriate search operation and then combines the search results using a sequence of reordering.
In particular, we designed the system so the {\ocsearch} operation has the priority: the result set contains all the images which match the given query with taking into account the classes drawn in the canvas (both object and colors), and not their spatial location. If the query includes also some scene tags (text box of the user interface), then the {\ansearch} is performed but only on the result set generated by the first {\ocsearch}. So in this case the {\ansearch} actually produce only a rescoring of the results obtained at the previous step. Finally, another rescore is performed using the {\bbsearch}.
If the user does not issue any annotation keyword in the interface, only the {\ocsearch} and {\bbsearch} are used. If, on the other hand, only one or more keywords are put in the interface, only the {\ansearch} is used to find the results. 

{\simsearch} is the only search operation that is stand-alone in our system, i.e. it is never combined with other search operations. However, we note that in future versions of {\oursystem} it may be interesting to also include the possibility of using {\simsearch} to reorder the results obtained from other search operations.

\section{Evaluation }\label{sec:exp}

\h{As already discussed in Sections \ref{sec:system_description} and \ref{sec4}, a} user query is executed as a combination of search operations ({\ansearch}, {{\bbsearch}}, {{\ocsearch}},  and {{\simsearch}}).  The final result set returned to the user highly depends on the results returned by each executed search operation. Each search operation is implemented in Apache Lucene using a specific \textit{ranker} (e.g. TF-IDF and BM25) that determines how the textual encoding of the database items are compared with the textual encoding of the query in order to provide the ranked list of results to the query.

In our first implementation of the system, used at the VBS competition in 2019, we tested for each search operation various rankers, and we estimated the performance of the system using our personal experience and feeling. 
Specifically, we tested a set queries with different rankers and we select the ranker that provided us with good results in the top positions of the returned items. However, given the lack of a ground truth, this qualitative analysis was based on a subjective feedback provided by a member of our team who explicitly looked at the top-returned images obtained with the various tested scenarios, and judged how good the results were. 

\hf{After the competition, we decided to have a more accurate approach to estimate the performance of the system, and the results of this analysis are discussed in this section.
As the choice of the rankers strongly influences the performance of the system, we decided to have a more in-depth and objective analysis based on this part of the system. The final scope, of this analysis, is finding for our system the best {\rankers} combination. Intuitively, the best combination of rankers is the one that, on average, puts more often good results (that is target results for the search challenge) at the top of the result list.
Specifically, we used the query logs acquired during the participation at the challenge. The logs store all the sequences of search operations that were executed as consequence of users interacting with the system.} By using these query logs, we were able to re-execute the same user sessions using different rankers. In this way we objectively measured the performance of the system, obtained when the same sequence of operation was executed with different rankers. 

We focus mainly on the rankers for the {{\bbsearch}}, {{\ocsearch}}, and {\ansearch}. We do not consider the {\simsearch} as it is an independent search operation in our system, and previous work \cite{AMATO2019IPM} already proved that the dot product (TF ranker) works well with the surrogate text encodings of the R-MAC descriptors, which are the features adopted in our system for the {\simsearch}.

\subsection{Experiment Design and Evaluation Methodology}

As anticipated before, our analysis makes use of the log of queries executed during the 2019 VBS competition. The competition was divided in three content search \textit{tasks}: visual Known-Item Search (\textit{visual KIS}), textual Known-Item Search (\textit{textual KIS}) and ad-hoc Video Search (\textit{AVS}), already described in Section \ref{sec:intro}. For each task, a series of \textit{runs} is executed. In each run, the users are requested to find one or more target videos. When the user believes that he/she has found the target video, he/she submits the result to the organization team, that evaluate the submission.

After the competition, the organizers of VBS provided us with the VBS2019 server dataset that contains all the tasks issued at the competition (target video/textual description, start/end times of target video for KIS tasks, and ground-truth segments for KIS tasks), the client logs for all the systems participating to the competition, and the submissions made by the various teams. We used the ground-truth segments and the log of the queries submitted to our system to evaluate the performance of our system under different settings. We restricted the analysis only to the logs related to textual and visual KIS tasks since ground-truths for AVS tasks were not available\footnote{Please note that for the AVS tasks the evaluation of the correctness of the results submitted by each team during the competition was made on site by members of a jury who evaluated the submitted images one by one. For these tasks, in fact, a predefined ground-truth is not available.}.

During the VBS competition a total of four users (two experts and two novices) interacted with our system to solve 23 tasks (15 visual KIS and 8 textual KIS). The total number of queries executed on our system for those tasks was 1600\footnote{We recall that, in our system, a new query is executed at each interaction of a user with the search interface and the images visualized in the browsing interface and their ranking is updated accordingly.}. 

In our analysis, we considered four different rankers to sort the results obtained by each search operation of our system. Specifically we tested the rankers based on the following text scoring function:
\begin{itemize}
    \item \emph{BM25}: Lucene's implementation of the well-known similarity function BM25 introduced in~\cite{Robertson94};
    \item \emph{TFIDF}: Lucene's implementation of the weighing scheme known as tfidf (Term Frequency-Inverse Document Frequency) introduced in~\cite{sparckjones1972}; 
    \item \emph{TF}: implementation of dot product similarity over the frequency terms vector;
    \item \emph{NormTF}: implementation of cosine similarity (the normalized dot product of the two weight vectors) over the frequency terms vectors.
\end{itemize}

Since we have three search operations and four \rankers, we have a total of 64 possible combinations. We denote each combination with a triplet {\triplet} where $R_{BB}$ is the ranker used for the {\bbsearch}, $R_{AN}$ is the ranker used for the {\ansearch}, and $R_{OC}$ is the ranker used for the {\ocsearch}.  
\h{In the implementation of VISIONE used at the 2019 VBS competition}, we employed the combination BM25-BM25-TF. With the analysis reported in this section, we compare all the different combinations in order to find the one that is most suited for the video search task.

For the analysis reported in this section we went trough the logs and automatically re-executed all the queries using the 64 different combinations of rankers in order to find the one that, with the highest probability, finds a relevant result (i.e. a keyframe in the ground-truth) in the top returned results. Each combination was obtained by selecting a specific ranker (among BM25, NormTF, TF, and TFIDF) for each search operation ({\bbsearch}, {\ansearch}, and {\ocsearch}).

\subsubsection{Evaluation Metrics}
\begin{figure}[!tb]
	\centering
		{\includegraphics[ trim= 0mm 0mm 0mm 0mm,clip,width=\columnwidth]{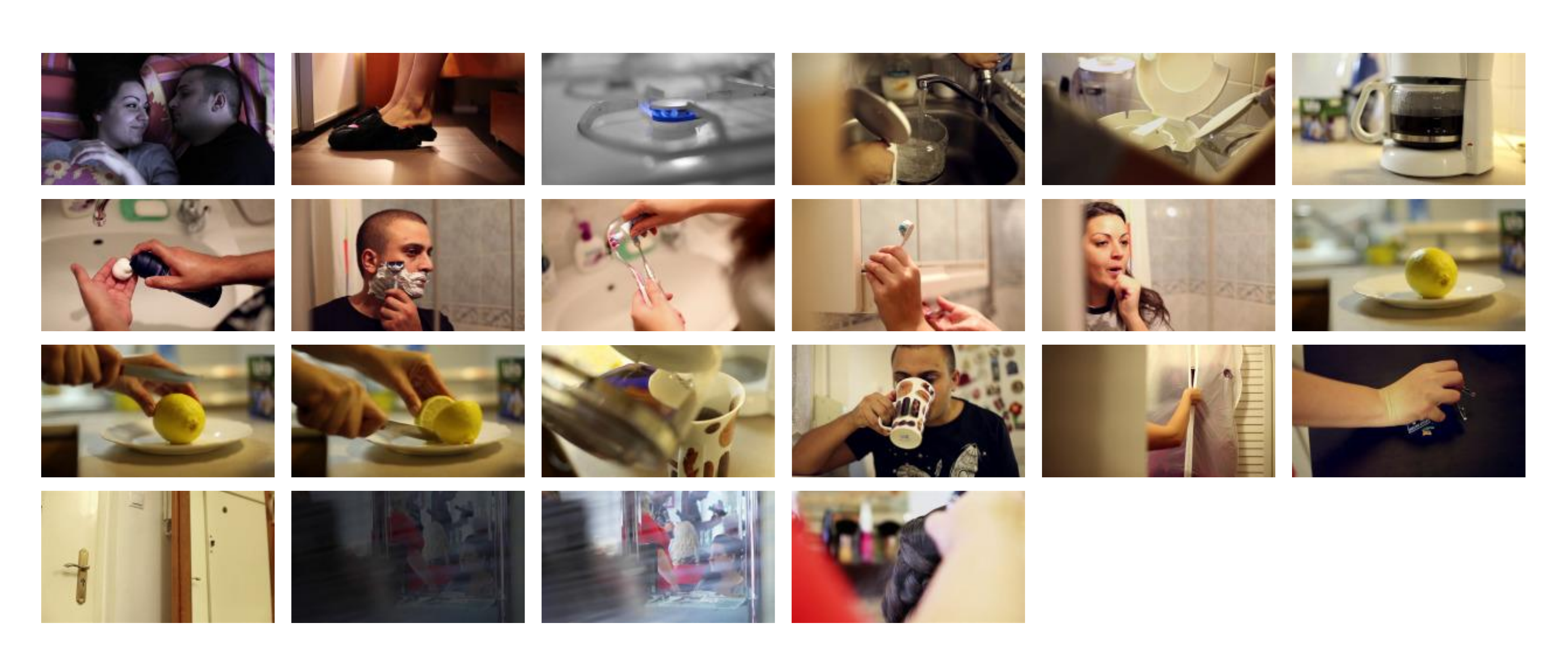}} %
	\caption[]{Example of the ground-truth keyframes for a 20 second video clip used as a KIS task at VBS2019. During the competition, our team correctly found the target video by formulating a query describing one of the keyframes depicting a lemon. However, note that most of the keyframes in the ground-truth were not relevant for the specific query submitted to our system. } 
 \label{fig:exampleGT}
\end{figure}
During the competition the user has to retrieve a video segment from the database using the functionalities of the system. A video segment is composed of various keyframes, which can be significantly different one from another, see Figure \ref{fig:exampleGT} as an example.

In our analysis, we assume that the user stops examining the ranked result list as soon as he/she finds one relevant result, that is one of the keyframes belonging to the target video. Therefore, given that relevant keyframes can be significantly different one from the other,  we do not take into account the rank position of \textit{all} the keyframes composing the ground-truth of a query, as required for performance measures like \textit{Mean Average Precision} or \textit{Discounted Cumulative Gain}. We want to measure how the system is good at proposing in the top position at least one of the target keyframes.

In this respect, we use the \textit{Mean Reciprocal Rank-MRR} (Equation \ref{eq:MRR}) as a quality measure, since it allows us to evaluate how good is the system in returning at least one relevant result (one of the keyframes of the target video) in top position of the result set.

Formally, given a set $Q$ of queries, for each $q \in Q$ \h{let $\{I^{(q)}_{1}, \dots, I^{(q)}_{n_q}\}$ the ground-truth, i.e., the set of ${n_q}$ keyframes of the target video-clip searched using the query $q$;} we define:
\begin{itemize}
    \item $rank(I^{(q)}_{j})$ as the rank of the image $I^{(q)}_{j}$ in the ranked results returned by our system after executing the query $q$
    \item $r_q=\min_{j=1, \dots n_q}rank(I^{(q)}_{j})$ as the rank of the first correct result in the ranked result list for the query $q$.
\end{itemize}
The Mean Reciprocal Rank (MRR) for the query set $Q$ is given by 
\begin{equation}
    MRR=  \dfrac{1}{|Q|}\sum_{q \in Q} RR(q),
\end{equation}\label{eq:MRR}
where the Reciprocal Rank (RR) for a single query $q$ is defined as
\begin{equation}
    RR(q)= \begin{cases}
    0 & \text{no relevant results} \\
    {1}/{r_{q}} & \text{otherwise}
    \end{cases}
\end{equation}

We evaluated the MRR for each different combination of rankers. Moreover, as we expect that a user inspects just a small portion of the results returned in the browsing interface, we also evaluate the performance of each combination in finding at least one correct result in the top $k$ positions of the result list ($k$ can be interpreted as the maximum number of images inspected by a user). To this scope we computed the MRR at position $k$ (MRR@k):
\begin{equation}
    MRR@k=  \dfrac{1}{|Q|} \sum_{q \in Q}  RR@k(q)
\end{equation}
where 
\begin{equation}
    RR@k(q)= 
    \begin{cases}
        0 & \text{$r_{q} > k$ OR no relevant results} \\
        {1}/r_{q} & \text{otherwise}
    \end{cases}
\end{equation}

In the experiments we consider values of $k$ smaller than $1000$, with a focus on values between $1$ and $100$ as we expect cases where a user inspects more than 100 results to be less realistic. 

\subsection{Results}
In our analysis, we used $|Q|=521$ queries (out of 1600 above mentioned) to calculate both $MRR$ and $MRR@k$. In fact the rest of the queries executed on our system during the VBS2019 competition are not eligible for our analysis since they are not informative to choose the best ranker configuration: 
\begin{itemize}
\item about 200 queries involved the execution of a {\simsearch}, a video summary or a filtering, whose results are independent of the rankers used in the three search operations considered in our analysis;
\item the search result sets of about 800 queries do not contain any correct result due to the lack of alignment between the text associated with the query and the text associated with images relevant to the target video. 
For those cases, the system is not able to display the relevant images in the result set regardless of the ranker used. In facts, the effect of using a specific ranker only affects the ordering of the results and not the actual selection of them.
\end{itemize}
\begin{figure}[t]
	\centering
		{\includegraphics[trim= 8mm 147mm 7mm 55mm,clip,width=\columnwidth]{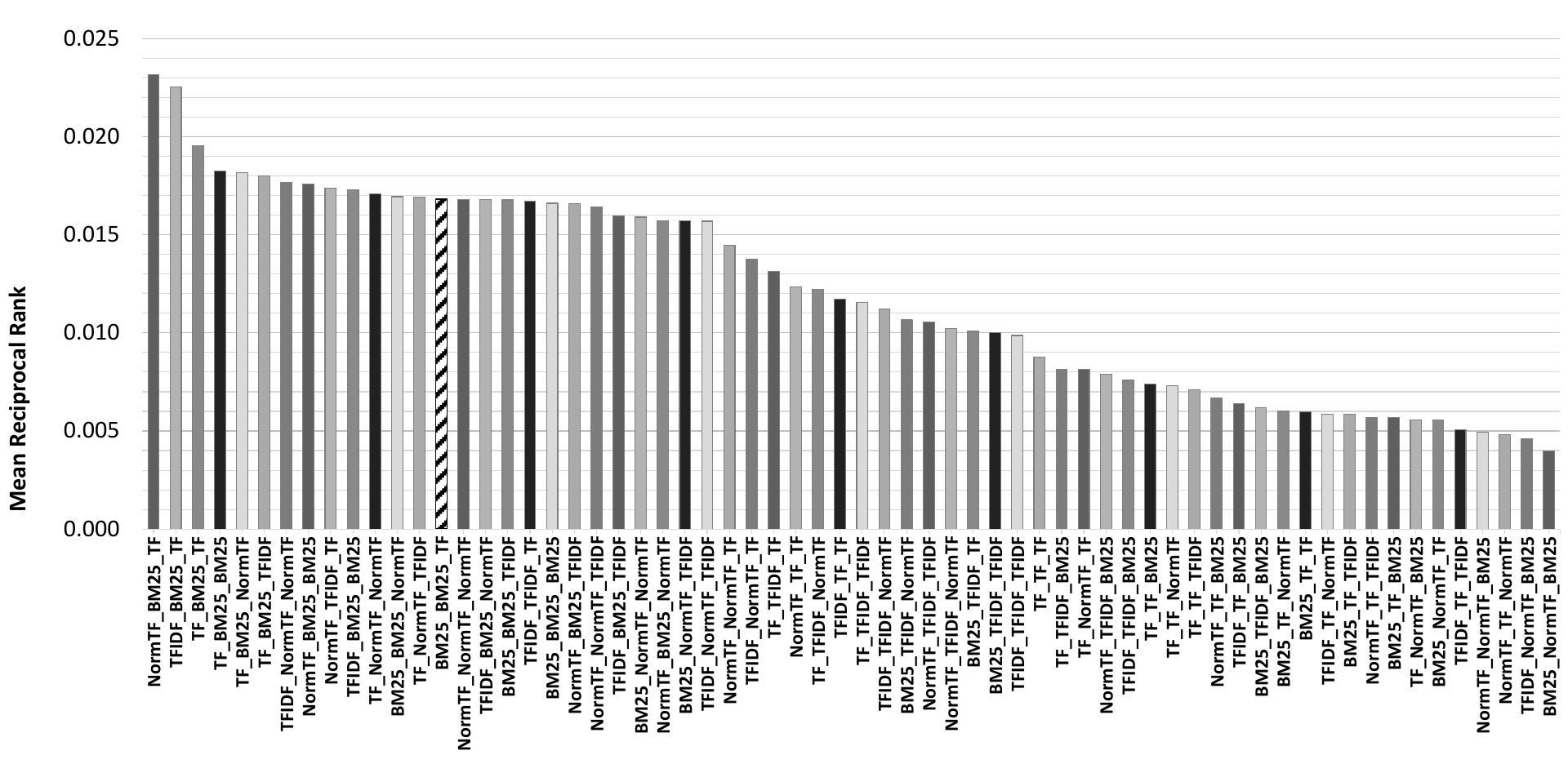}} 
	\caption[]{Mean Reciprocal Rank of the 64 combinations of ranker: the 
	one filled with diagonal lines in the graph 
	is the combination used at the VBS2019 competition.} 
 \label{fig:MRR}
\end{figure}
Figure~\ref{fig:MRR}, which reports the MRR values of all 64 combinations, shows that there is a significant difference between the best and the worse combination. Note that the combination that we used at VBS 2019 (indicated with diagonal lines in the graph), and that was chosen according to subjective feelings, has a good performance, but it is not the best. In fact, we noticed that there exist some patterns in the combinations of the rankers used for the {\ocsearch} and the {\ansearch} which are particularly effective and some which, instead, provide us with very poor results. For example, the combinations that use \textit{TF} for the {\ocsearch} and \textit{BM25} for the {\ansearch} gave us the overall best results. While the combinations that use \textit{BM25} for the {\ocsearch} and the \textit{NormTF} for the {\ansearch} have the worse performance. Specifically, we have an MRR of $0.023$ for the best (NormTF-BM25-TF) and $0.004$ for the worse (BM25-NormTF-BM25).
\begin{figure}[tp]
	\centering
		\includegraphics[ trim= 20mm 85mm 20mm 90mm,clip,width=0.568\columnwidth]{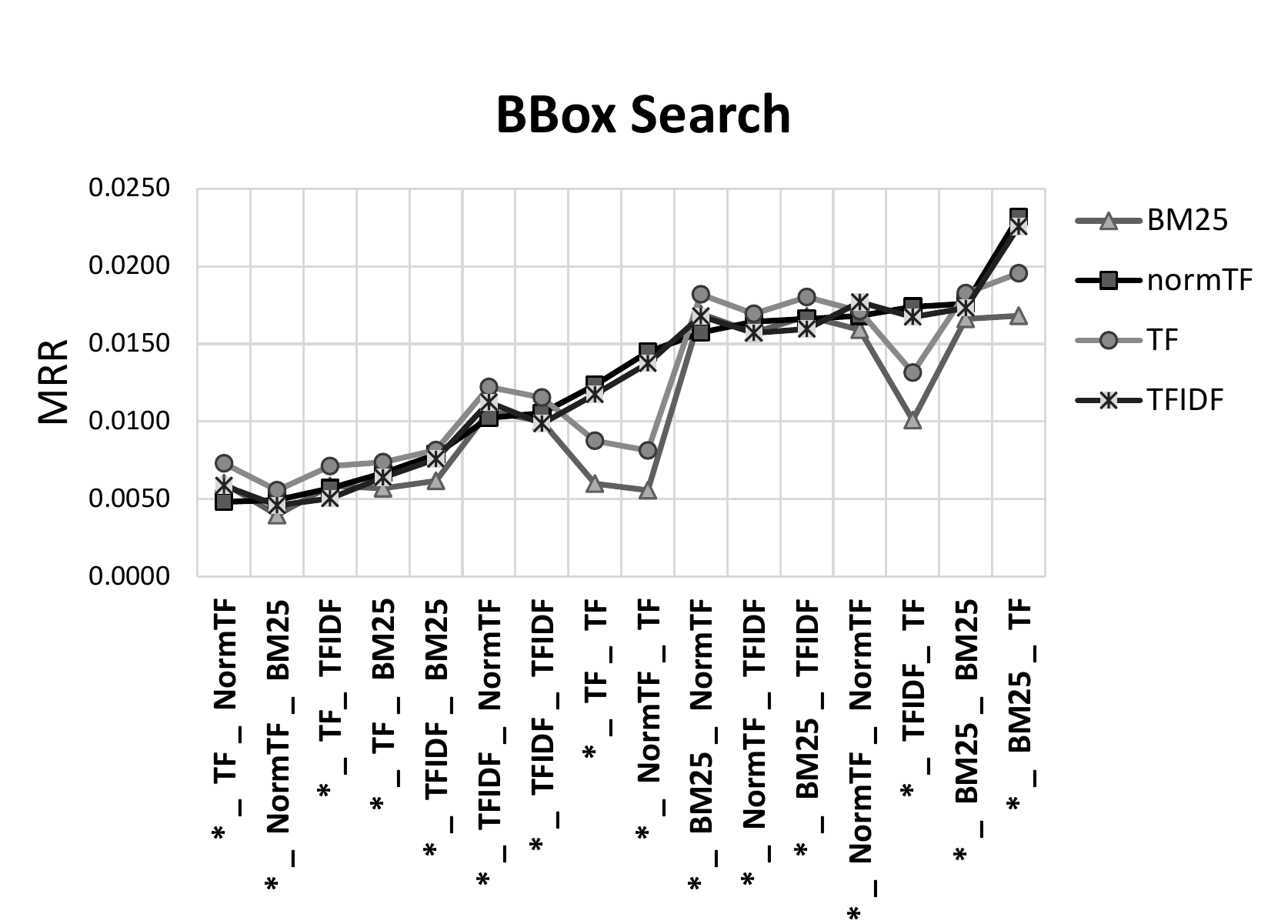} 	
		\includegraphics[ trim= 20mm 85mm 20mm 90mm,clip,width=0.568\columnwidth]{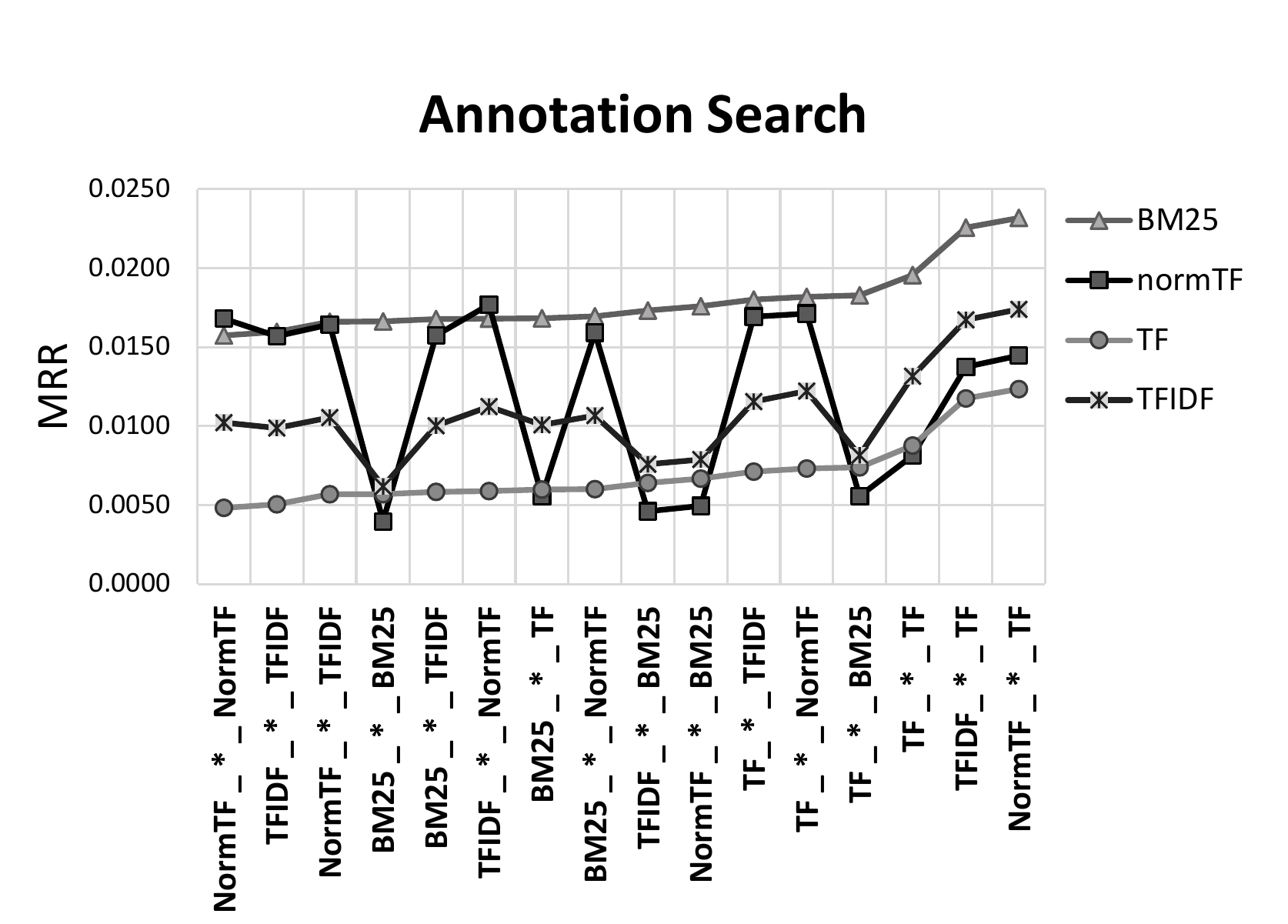} 
		\includegraphics[ trim= 20mm 85mm 20mm 90mm,clip,width=0.568\columnwidth]{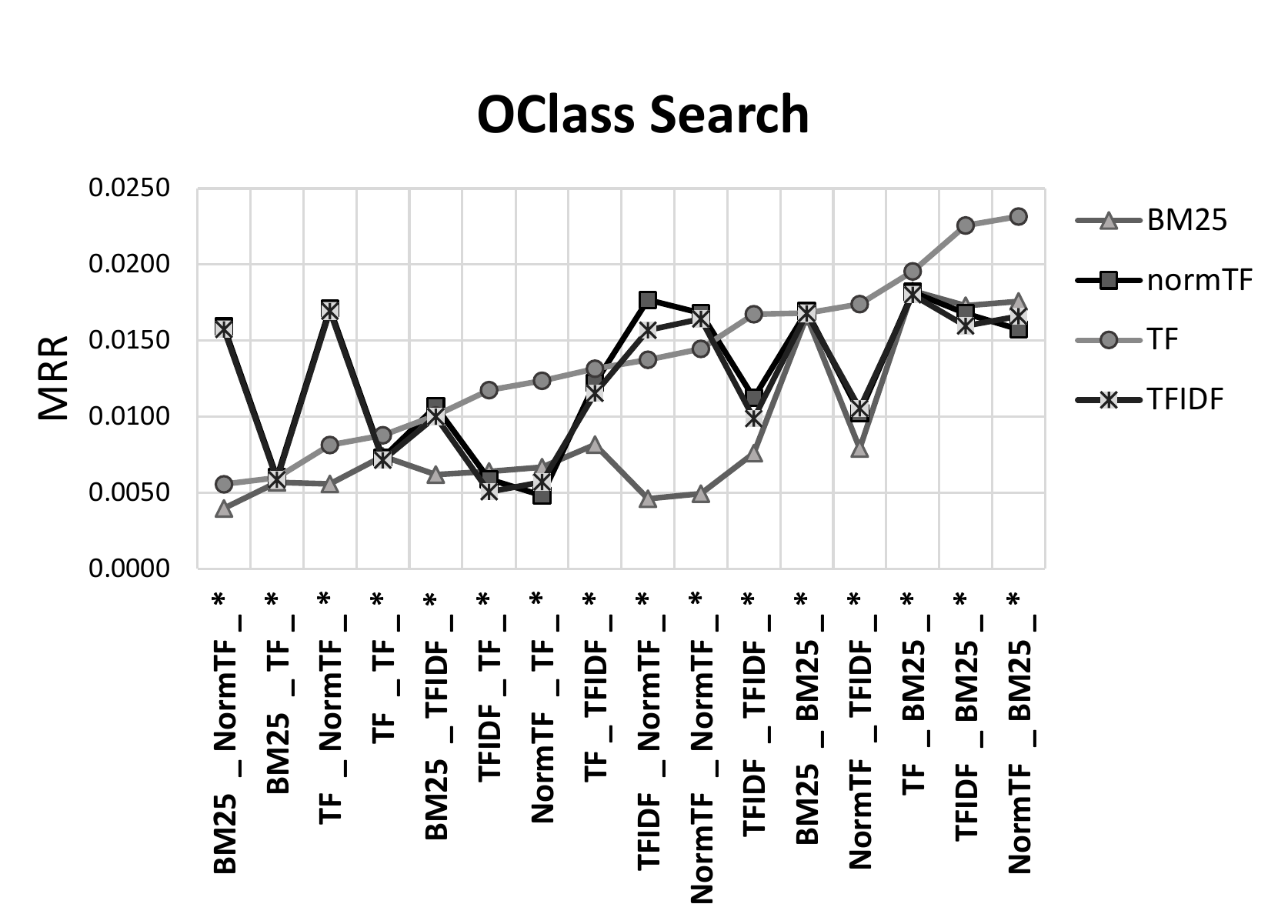} %
	\caption[]{Mean Reciprocal Ranks varying the ranker for each search operation. Each configuration is denoted with a triplet {\triplet} where $R_{BB}$ is the ranker used for the {\bbsearch}, $R_{AN}$ is the ranker used for the {\ansearch}, and $R_{OC}$ is the ranker used for the {\ocsearch}.}
 \label{fig:MRR-Ranker}
\end{figure}

From a further analysis of the MRR results, it turned out quite clearly that for the {\ansearch} the ranker \textit{BM25} is particularly effective, while the use of the \textit{TF} ranker highly degrades the performance. This is even more evident in the results shown in Figure~\ref{fig:MRR-Ranker}, where for each search operation we expose the MRR values obtained for a fixed ranker while varying the rankers used with the other search operations. 
It also turned out that for the {\bbsearch} the \textit{TFIDF} and the \textit{NormTF} have better performance than \textit{TF} and \textit{BM25} rankers. Moreover, for the {\ocsearch} the \textit{BM25} has the worse performance in general and the \textit{TF} ranker is the one that provides us with the best results.
\begin{figure}[tp]
	\centering
		{\includegraphics[trim= 20mm 100mm 20mm 100mm,clip,width=0.935\columnwidth]{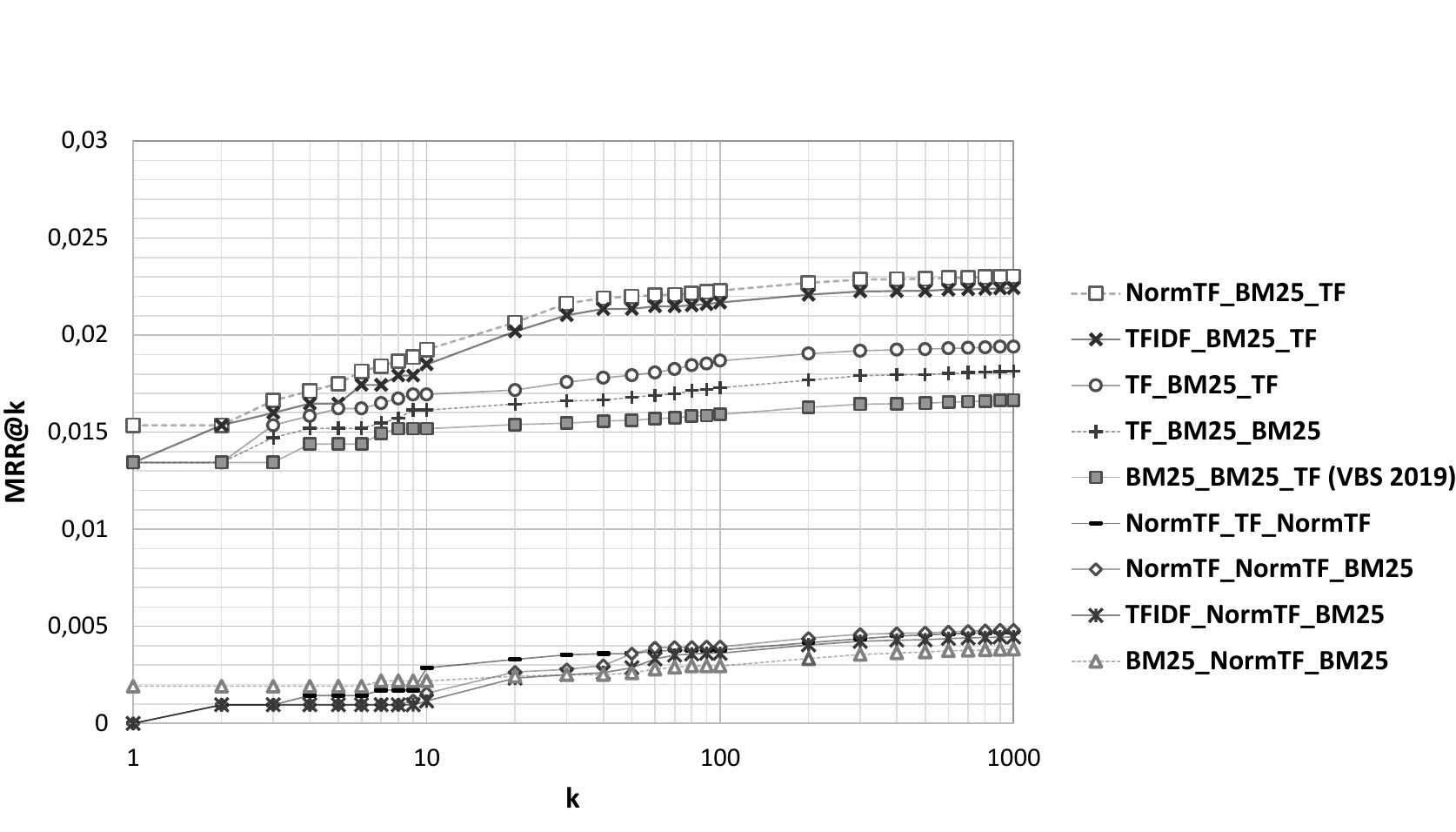}} %
    	\caption[]{MRR@k for eight combinations of the rankers (the four best, the four worst and the setting used at VBS2019) varying $k$ from 1 to 1000.} 
 \label{fig:MRRatK}
\end{figure}

Furthermore, to complete the analysis on the performance of the rankers, we analyze the MMR@k, where $k$ is the parameter that controls how many results are shown to the user in the results set. The results are reported in Figure~\ref{fig:MRRatK}, where we varied $k$ between $1$ and $1,000$. In this case, for a better understanding of the chart we reported only eight combinations (the four with the best MMR@k, the four with worst MMR@k, and the configuration used at VBS2019). 
In conclusion, we identified the configuration \textit{NormTF-BM25-TF} as the best one for the {\triplet} search operations which provides as a relative improvement of $38\%$ in $MRR$ and $40\%$ in $MRR@100$ with respect to the setting previously used at the VBS competition.

\hf{
\subsection{Efficiency and Scalability Issues}
As we stated in the introduction, the fact that the retrieval system proposed in this article is built on top of a text search engine guarantees in principle efficiency and scalability of queries. This has been practically verified by obtaining average response times of less than a second for all types of queries (even more complex ones). On the scalability of the system, we can make some optimistic assumptions because we have not conducted experiments on it. 
This optimistic assumption is based on the observation that if the ``synthetic'' documents generated for visual search by similarity, and for the localization of objects and colors behave as textual documents then the scalability of our system is comparable to that of commercial Web search engines. 
To this end, with regard to the scalability of visual similarity as we rely on the technique used to index R-MAC descriptors based on scalar quantization, the reader is referred to the work \cite{AMATO2019IPM}, in which the scalability of this approach is proven. On the other hand, as far as objects and colors are concerned, we have analyzed the sparsity of the inverted index corresponding to synthetic documents and we have seen that it is around 99.78\%.  Moreover, since the queries are similar in length to those of natural language search scenarios (i.e. they have few terms), the scalability of the system is guaranteed at least as much as that of full-text search engine scenarios.
}
\section{Conclusions}\label{sec:conclusion}
In this paper, we described a frame-based interactive video retrieval system, named {\oursystem}, that participated to the Video Browser Showdown (VBS) contest in 2019. {\oursystem} includes several retrieval modules and supports complex multi-modal queries, including query by keywords (tags), query by object/color location, and query by visual example. A demo of {\oursystem} running on the VBS V3C1 dataset is publicly available at \url{http://visione.isti.cnr.it/}.

{\oursystem} exploits a combination of artificial intelligence techniques to automatically analyze the visual content of the video keyframes and extract annotations (tags), information on objects and colors appearing in the keyframes (including the spatial relationship among them), and deep visual descriptors. 
A distinct aspect of our system is that all these extracted features are converted into specifically designed text encodings that are then indexed using a full-text search engine. The main advantage of this approach is that {\oursystem} can exploit the latest search engine technologies, which today guarantee high efficiency and scalability.

The evaluation reported in this work shows that the effectiveness of the retrieval is highly influenced by the text scoring function (ranker) used to compare the textual encodings of the video features. In fact, by performing an extensive evaluation of the system under several combinations, we observed that an optimal choice of the \h{ranker} used to sort the search results can improve the performance in terms of Mean Reciprocal Rank up to an order of magnitude. Specifically, for our system we found out that \textit{TF}, \textit{NormTF}, and \textit{BM25}, are particularly effective for comparing textual representations of object/color classes, object/color bounding boxes, and tags, respectively.

	\section*{Acknowledgements}
	This work was partially funded by ``Smart News: Social sensing for breaking news", CUP CIPE D58C15000270008, by VISECH ARCO-CNR, CUP B56J17001330004, and by ``Automatic Data and documents Analysis to enhance human-based processes" (ADA), CUP CIPE D55F17000290009. We gratefully acknowledge the support of NVIDIA Corporation with the donation of the Tesla K40 GPU used for this research.

	\bibliographystyle{unsrt}

	\bibliography{bib}

\end{document}